\documentclass{article}

\usepackage{arxiv}

\usepackage[utf8]{inputenc} 
\usepackage[T1]{fontenc}    
\usepackage{hyperref}       
\usepackage{url}            
\usepackage{booktabs}       
\usepackage{amsfonts}       
\usepackage{nicefrac}       
\usepackage{microtype}      
\usepackage{lipsum}

\usepackage{amsmath,graphicx}
\usepackage{amssymb}
\usepackage{extarrows}
\usepackage{color}
\usepackage{mathtools}
\usepackage{lipsum,cuted}
\usepackage{algorithmic}
\usepackage{cleveref}
\usepackage[algoruled,boxed,linesnumbered]{algorithm2e}
\usepackage[shortlabels]{enumitem}
\usepackage{booktabs} 
\usepackage{multirow}

\DeclareMathOperator{\E}{\mathbb{E}}
\def\*#1{\mathbf{#1}}

\definecolor{brown}{rgb}{0.59, 0.29, 0.0}
\newcommand{\bLozenge}{\color{brown}{\mathbin{\blacklozenge}}}

\title{Multimodal Representation Learning using Deep Multiset Canonical Correlation Analysis}

\author{
  Krishna Somandepalli, Naveen Kumar, Ruchir Travadi, Shrikanth Narayanan \\
  Signal Analysis and Interpretation Laboratory, University of Southern California\\
}

\begin{document}
\maketitle

\begin{abstract}
We propose Deep Multiset Canonical Correlation Analysis (dMCCA) as an extension to representation learning using CCA when the underlying signal is observed across multiple (more than two) modalities. 
We use deep learning framework to learn non-linear transformations from different modalities to a shared subspace such that the representations maximize the ratio of between- and within-modality covariance of the observations. 
Unlike linear discriminant analysis, we do not need class information to learn these representations, and we show that this model can be trained for complex data using mini-batches.
Using synthetic data experiments, we show that dMCCA can effectively recover the common signal across the different modalities corrupted by multiplicative and additive noise. We also analyze the sensitivity of our model to recover the correlated components with respect to mini-batch size and dimension of the embeddings.
Performance evaluation on noisy handwritten datasets shows that our model outperforms other CCA-based approaches and is comparable to deep neural network models trained end-to-end on this dataset.
\end{abstract}
\vspace{-0.1in}
\section{Introduction and Background}
\label{sec:intro}
In many signal processing applications, we often observe the underlying phenomenon through different modes of signal measurement, or \textit{views} or \textit{modalities}. 
For example, the act of running can be measured by electrodermal activity, heart rate monitoring and respiratory rate. We refer to these measurements as \textit{modalities}. These observations are often corrupted by sensor noise, as well as the variability of the signal across people, influenced by interactions with other physiological processes.
The question then becomes: Given that these modalities are observing the same phenomenon, how do we learn to represent the information that is common to them. \textit{Multimodal representation learning} often refers to this task of learning a subspace that captures the information shared among modalities.

Multimodal representations have been shown to capture the variability of the underlying signal better than a single modality \cite{sridharan2008information}, and generalize better to unseen data.
Recently, deep neural network (DNN) methods in this domain have been successful for downstream tasks such as clustering or classification. Some examples of related work are multimodal autoencoders \cite{ngiam2011multimodal}, RBM \cite{srivastava2012multimodal}, and our work on autoencoder shared representations \cite{somandepalli2018multimodal}. Most of these methods optimize some form of a reconstruction error. In this work we are interested in extending deep CCA to multiple modalities.

CCA \cite{hotelling1936relations, anderson1958introduction} and kernel CCA \cite{akaho2006kernel} can find projections of two modalities to maximially correlate them.
They have been applied for many applications--from unsupervised learning of fMRI data \cite{hardoon2004canonical} to decreasing sample complexity for regression \cite{kakade2007multi}.
A drawback of CCA is that it can only learn a \textit{linear} projection that maximizes the correlation between modalities. 
This can be somewhat alleviated by KCCA, which uses a kernel to learn nonlinear representations. 
But, the ability of KCCA to generalize is limited by the fixed kernel.

Deep CCA (DCCA \cite{andrew2013deep}) addresses this limitation by learning non-linear transformations of the data that maximize the correlation between them. 
DCCA has been successfully applied to several applications for high dimensional data with complex relationships.
Examples include unsupervised learning of acoustic features from articulatory data \cite{wang2015unsupervised}, matching images with caption \cite{yan2015deep} and multilingual word-embeddings \cite{lu2015deep} 
By definition, CCA and DCCA can only work with two modalitites and do not scale for multiple modalities.

One of the first works to extend CCA for several `sets of variables' was proposed in \cite{kettenring1971canonical}. 
Subsequently, the seminal work by Nielsen \cite{nielsen2002multiset} examined different formulations of \textit{multiset} CCA, and its applications.
Another method for multiset CCA is generalized CCA (GCCA \cite{horst1961generalized}).
It has been successfully applied to multi-set problems such as multi-view latent space \cite{sharma2012generalized} and latent semantic analysis \cite{rastogi2015multiview}.
Deep GCCA \cite{benton2017deep} has been proposed to generalize GCCA. However, (D)GCCA only considers within-modality covariances, similar to principal component analysis (PCA), i.e, it captures components with maximum variance within a modality. 

Complementary to GCCA, multiset CCA \cite{mckeon1967canonical, kettenring1971canonical} considers both between- and within-modality variances. 
A formal characterization of MCCA was published in \cite{parra2018correlated} to show that the MCCA solution is akin to maximizing the `reliability' of repeated measurements.
It is important to note that the ratio of covariances formulation here is similar to the ratio of scatter matrices used in linear discriminant analysis (LDA). 
However LDA and MCCA are conceptually different in that LDA needs class information but MCCA only needs congruent modalites.
The relation between LDA and MCCA is detailed in \cite{parra2018correlated}.
However, MCCA is still limited to learning only linear projections. In this work, we propose deep MCCA (dMCCA) that enables us to learn non-linear transformations. We also show that dMCCA can be trained with stochastic methods, making it possible to scale the method for complex and large datasets.

\textbf{CCA, GCCA and deep learning variants: }
\newline
Let $\ (\*X^1, \*X^2)$ be data matrices of two modalities, with covariances, $\ (\*{\Sigma}_{11},\*{\Sigma}_{22})$ and cross covariance $\ \*{\Sigma}_{12}$. 
CCA finds linear projections in the direction that maximizes the correlation coefficient.
In DCCA, the top-most layers of two DNNs are used instead of the data matrices, i.e., $\ \*X^{l} \rightarrow \*H^{l}$ to estimate the covariances,$\ \*{\Hat{\Sigma}}$.
The loss function to optimize this network is the sum of the top \textit{k} singular values of the matrix $\ \*T =  \*{\Hat{\Sigma}}_{11}^{-1/2}\*{\Hat{\Sigma}}_{12}\*{\Hat{\Sigma}}_{22}^{-1/2}$. DCCA can be optimized full-batch \cite{andrew2013deep}, or with stochastic methods \cite{yan2015deep}.\newline
\textbf{GCCA} works with$\ T\times D\times N$ tensor (notation consistent with \cite{parra2018correlated}) where $\ T$ is the number of data samples of dimension$\ D$ for $\ N>2$ modalities.
Let $\ \*X^l\in\mathbb{R}^{T\times D}, l=1,...,N$ where the feature dimension$\ d_l$ can vary with modality. 
GCCA finds a low-dimensional, $\ K<D$ orthonormal space such that its Frobenius norm to the input signal is minimized for all modalities.
Its deep variant, DGCCA extends this formulation to DNN by maximizing the top \textit{K} eigenvalues of the sum of projection matrices that whitens all$\ \*H^l, l\in[1,...,N]$ matrices. See \cite{benton2017deep} for more details.\newline
\vspace{-0.1in}
\section{Multiset CCA and dMCCA}\label{sec:dmcca}
In this section, we describe our formulation of dMCCA and give a sketch to derive gradients for backpropagation following \cite{andrew2013deep,dorfer2015deep}.
Let $\ \*X^l \in \mathbb{R}^{T\times D}, l=1,....N$ be the data matrices of \textit{N} modalities, the the inter-set correlation (ISC ,\cite{parra2018correlated}) is defined as:
{\small
\begin{eqnarray}
    \rho_d = \frac{1}{N-1}\frac{\*v_d^{\top}\*R_B\*v_d}{\*v_d^{\top}\*R_W\*v_d} ,\quad d=1,..,D
\end{eqnarray}}
where $\ \*R_B\text{ and }\*R_W$ referred to as the \textit{between-set} and \textit{within-set} covariance matrices defined as:
{\small
\begin{eqnarray}\label{eqn:rb}
    \*R_B = \Sigma_{l=1}^{N}\Sigma_{k=1,k\neq l}^{N}\Bar{\*X}^{l}(\Bar{\*X}^{k})^{\top} \\
    \*R_W = \Sigma_{l=1}^{N}\Bar{\*X}^{l}(\Bar{\*X}^{l})^{\top}
\end{eqnarray}}
where $\ \Bar{\*X}=\*X-\E(\*X)$ are the centered datasets. We omit the common scaling factor$\ (T-1)^{-1}N^{-1}$ here. 
MCCA finds the projection vectors$\ \*v_d, d=1,...,D$ to maximize the ratio of sums of between-set and within-set covariances by solving this generalized eigenvalue (GEV) problem:
{\small
\begin{eqnarray}\label{eqn:gev}
    \*R_B\*V = \*R_W\*V\*{\Lambda} ,\quad \text{diagonal matrix},\quad \*{\Lambda}_{dd} = \rho_d 
\end{eqnarray}}
In summary, MCCA finds the projection of the data that maximizes ISC by finding the principal eigenvectors of between-set over within-set covariance. In simpler terms, MCCA examines the eigenvalue decomposition of $\ \*R_W^{-1}\*R_B$ when $\ \*R_W$ is invertible.

In the dMCCA formulation,$\ \*X^{\cdot}\rightarrow\*H^{\cdot}$. With increasing \textit{N}, computing $\ \*R_B$ can be computationally expensive because all pairs of modalities need to be considered. 
Instead we estimate the \textit{total covariance}$\ \*R_T$, and then $\ \*R_B = \*R_T-\*R_W$ as:
{\small
\begin{eqnarray}
    \*R_T = N^2(\Bar{\*H}^*-\*{\mu}\*1^{\top})(\Bar{\*H}^*-\*{\mu}\mathbf{1}^{\top})^{\top}
\end{eqnarray}}
where $\ \*H^* = \Sigma_{l=1}^{N}\*H^{l}$ is the average of all modalities, and $\ \*{\mu}=\frac{1}{T}\Sigma_{t=1}^{T}(\frac{1}{N}\Sigma_{l=1}^{N}\*h_t^l)$ -- the mean of all samples and all modalities.

The pseudo code of the our algorithm is given in \textbf{Alg. \ref{alg1}}.
{\tiny
\SetKwInput{kwInit}{Input}
\SetKwInput{kwInitb}{Output}
\SetKwInput{kwInitc}{Initialize}
\begin{algorithm}[t!]\label{alg1}
\caption{dMCCA}
\SetAlgoLined
\kwInit{\textit{N}-inputs with batch size \textit{M} $\ [\*X^1,...,\*X^N] , \*X^l\in\mathbf{R}^{M\times D}\text{ learning rate }\eta$}
\kwInitb{\textit{K}-dim representations from a \textit{d}-layer network $\ [\*H^1_d,...,\*H^N_d]$}
\kwInitc{\textit{N}-network weights $\ [\*W^1,...,\*W^N]$}
\While{not converged}{
\For{l=1,2,..,N}{
$\ \*H^l \gets$ forward pass of $\ \*X^l$ with $\ \*W^l$
}
 Estimate $\ \*R_B$ and $\ \*R_W$ for $\ \*H^l, l=1,...,N$\\
 Solve $\ \*V$ in (\ref{eqn:gev}) by factorizing $\ \*R_W=\*L\*L^{\top}$\\
 Compute $\ \mathcal{L} = \frac{\text{Tr}(\*V^{\top}\*R_B\*V)}{\text{Tr}(\*V^{\top}\*R_W\*V)}=\frac{1}{D}\Sigma_{d=1}^D\rho_d$\\
\For{l=1,2,..,N}{
$\ \Delta\*W^l\gets\text{ backprop}(\partial\mathcal{L}/\partial\*H^l, \*W^l)$\\
$\ \*W^l\gets\*W^l - \eta\Delta\*W^l$
}
}\end{algorithm}
}
We initialize the weights of the \textit{N} d-layer networks to perform a forward pass of the input data and obtain the top-most layer activations, $\ \*H^l$. 
We then estimate between-set and within-set covariances (eqn. 2-3) and solve the GEV problem (eqn. 4) using Cholesky decomposition of $\ \*R_W$ (line 6 Alg. 1) as described in \cite{parra2018correlated}.
Our objective is to maximize the average ISC (eqn 3) which is recomputed with the eigenvectors $\ \*V$ (line 7 Alg. 1). 
Notice that the GEV solution$\ \*V$ simultaneously diagnolizes $\ \*R_B\text{ and }\*R_W$, hence the loss function is equivalent to the ratio of the diagonal elements.
The key is to begin with the GEV problem (eqn 6):
{\small
\begin{eqnarray}
  \rho_d\*R_W\*v_d = \*R_B\*v_d = (\*R_T-\*R_W)v_d \\
  \implies \*R_T\*v_d = (\rho_d+1)\*R_W\*v_d
\end{eqnarray}}
The partial derivative of the eigenvalue$\ \rho_d$ with respect to hidden layer representations$\ \*H^l$ can be written as follows \cite{de2011derivatives}:
{\small
\begin{eqnarray}
\frac{\partial\rho_d}{\partial\*H^l} = \*v_d^{\top}\bigg(\frac{\partial\*R_T}{\partial\*H^l} - (\rho_d+1)\frac{\partial\*R_W}{\partial\*H^l}\bigg)\*v_d
\end{eqnarray}}

WLOG, assume $\ \*{\mu}=0$ in eqn (5) and with simpler notation $\ \*H^*\rightarrow\*H$ the partial derivative of $\ \*R_T$ in a mini-batch size of$\ M$ is:
{\small
\begin{eqnarray}
    \frac{\partial[\*R_T]_{ab}}{\partial[\*H]_{ij}} = 
    \begin{cases}
        \frac{2}{M-1}[\*H]_{ij} - \frac{1}{M}\Sigma_k[\*H]_{ik}\quad \text{if } a=b=i \\
        \frac{1}{M-1}[\*H]_{bj} - \frac{1}{M}\Sigma_k[\*H]_{bk}\quad \text{if } a=i,b\neq i \\
        \frac{1}{M-1}[\*H]_{aj} - \frac{1}{M}\Sigma_k[\*H]_{ak}\quad \text{if } a\neq i,b=i \\
        0 \quad \text{ else}
    \end{cases}
\end{eqnarray}}
To derive gradients of $\ \*R_W$, we refer the reader to \cite{andrew2013deep}. 

\section{Experiments}
\label{sec:exp}

\subsection{Synthetic data}
In order to test if dMCCA is learning highly correlated components, we generate synthetic observations similar to that in \cite{parra2018correlated} where the number of common signal components across the different modalities is known apriori. Because the source signal is given, we can build a supervised deep learning model to reconstruct the source signal. This provides an empirical upper-bound on performance.\newline
\textbf{Data Generation: } 
Consider$\ T$ signal and noise components for$\ N$ modalities $\ \*s^l_t\in\mathbb{R}^K\text{ and }\*n^l_t\in\mathbb{R}^D\, t=1,...,T\,, l=1,...,N\,, K<D$, drawn from standard normal distribution. 
Because our objective is to obtain correlated components across the modalities, the signal component is same across all$\ N$ modalitites, i.e, $\ \*s^l_t \approx \*s_t$, but corrupted with a modality-specific noise$\ \*{\eta^l}$.  
Thus, the signals were mapped to the measurement space as$\ \*x_{s,t}^l=\*A_s^l\*s_t + \*{\eta}^l, \*x_{n,t}^l=\*A_n^l\*n^l_t$. 
The multiplicative noise matrices were generated as$\ \*A_s^l = \*O_s^l\*D_s^l \in \mathbb{R}^{D\times K}\text{ and }\*A_n^l=\*O_n^l\*D_n^l \in \mathbb{R}^{D\times D}$. The matrices$\ \*O_s^l\in\mathbb{R}^{D\times K}\text{ and }\*O_n^l\in\mathbb{R}^{D\times D}$ have random orthonormal columns. 

Unlike \cite{parra2018correlated}, we used different matrices$\ \*A_s^l\text{ and }\*A_n^l$ to simulate a case where the different views of the underlying signal are corrupted by different noise. 
As is the case with many real world datasets, the noise in the measurement signal is further spatially correlated. 
We simulated this by $\ \*x_{n,t}^l\gets \alpha\*x_{n,t}^l + (1-\alpha)\*x_{n,t-1}^l , \alpha \in [0,1]$. 
Finally the SNR of the measurements was controlled by $\ \beta$ to generate the observation data as $\ \*y^l_t = \beta\*x_{s,t}^l + (1-\beta)\*x_{n,t}^l , \beta \in [0,1]$ resulting in a data matrix of size $\ T\times\ D\times N$.
For all our experiments, we generated data with $\ T=100000, D=1024, K=10, N=5, \beta = 0.7\text{ and  }\alpha=0.5$. Detailed analysis by varying these parameters is part of our future work.

\textbf{Supervised learning baseline: } We trained $\ N$ neural networks to obtain an empirical upper-bound on the performance of estimating the signal components. The input to these networks was $\ [\*X^1,...,\*X^N]$ and they were supervised to reconstruct the signal $\ [\*s^1,...,\*s^N]$.
In our experiments, with $\ N=5$ we used five identical networks, each with an input layer of 1024 nodes, followed by 512 nodes and 10 nodes (to match the number of correlated components).
We refer to this DNN as \textbf{Supervised-512-10}. 
It was trained to minimize the mean squared error (MSE) between the outputs and the signal components. 
Additionally, we also varied the activation functions (linear, tanh) and mini-batch sizes. 
A dropout of 0.2 was included for the networks with tanh activation to prevent overfitting. 
All models were trained using SGD with Nesterov momentum of 0.9 (learning rate=1e-3) with with 20\% of the data for validation (val) and 10\% for testing.
\vspace{-0.1in}
\subsubsection{dMCCA for synthetic data}\label{sec:dmcca-synth}
The dMCCA network architecture is similar to the aforementioned model with two modifications: 1) There was no supervision, instead the mean ISC between the modalities was maximized (See \textbf{Algo 1}), and 2) number of dimensions of the output embedding, i.e., the number of correlated components was varied as $\ K=[5, 10, 15, 20, 40, 50, 64, 128]$. We denote this model as \textbf{dMCCA-512-K}. 
The output layer dimension \textit{K} was varied to examine its effect on the similarity of the DNN representations to the source signal. 
This is important, since in real world applications the number of correlated components is not known apriori. 
In our preliminary experiments we noted that RMSProp (learning rate = 1e-3, decay = 0.9) instead of SGD with momentum yielded stable results in terms of loss at each epoch for the train and val data. 
Early stopping criterion was employed for both the supervised and dMCCA models (i.e, stop training when validation loss is less than 1e-6 for at least 5 consecutive epochs). All experiments were repeated ten times with different train-val-test partitions of the synthetic data.
\vspace{-0.1in}
\subsection{Noisy-MNIST (n-MNIST) experiments}
We used the n-MNIST data \cite{basu2017learning} to demonstrate that dMCCA algorithm can be used to learn the correlated signal components using observations from different modalities. n-MNIST was created using the handwritten digits dataset, MNIST \cite{lecun1998mnist} by adding 1) additive white Gaussian noise (AWGN) 2) motion blur and, 3) reduced contrast + AWGN. The noise parameters are described in \cite{basu2017learning}. In all experiments, we used 50000 samples for training, 10000 for validation and 10000 for testing. The ten digit classes in the dataset were nearly balanced, hence we only report accuracy.

\textbf{dMCCA for n-MNIST: }
The network architecture here was similar to the dMCCA model for synthetic data but with three branches. Each branch had an input layer of 784 nodes, followed by 1024 and \textit{K} nodes , i.e, \textbf{dMCCA-1024-K} . 
We used the val set for early stopping. 
The trained model was then applied on the test partition. 
Representations from the three modalities were concatenated as input features for SVM. 
The SVM parameters were tuned on the val-set using grid-search and accuracy from a 10-fold cross-validation (CV) on the test set was reported. 
All experiments were repeated by varying mini-batch size and embedding dimension. We also used k-means clustering algorithm to assess linear separability of the ten classes. 

\textbf{Baseline experiments: }
We setup five baseline experiments to compare the performance of dMCCA for classification. For all experiments, we report the 10-fold CV accuracy on the test-set, after tuning the system (SVM or DNN) parameters on the val-set. We also varied the feature or embedding size,$\ K$ and the mini-batch sizes (See Fig. 2 for ranges) where applicable. We used McNemar's $\ \chi^2$ test to statistically compare performance.

\textbf{1. Supervised baseline}:
The architecture here was similar to \textbf{Supervised-512-10}, except with three branches: 784 dim input followed by 1024 and \textit{K} nodes. The three \textit{K}-dimensional top-most layers were concatenated and input to a softmax layer of 10 nodes for classifying the 10 digits. 
The concatenated layer ensures that the information across the three modalities is shared in this embedding. The model was optimized with categorical cross-entropy using early stopping criteria as described before.
Because the model is supervised end-to-end, this is a very competitive baseline to compare with dMCCA.
Arguably it may not be the best architecture for classifying n-MNIST dataset, but our objective here was to obtain an upper-bound in DNN performance using end-to-end training.\newline
\textbf{2. PCA:}
We compare with PCA because like CCA, it performs dimensionality reduction. We concatenated the 784-dim vectors of three modalities of n-MINST and perform PCA on test-set by estimating the covariance on the train set. $\ K$ principal components are input as features to SVM.\newline
\textbf{3. DCCA:}
We chose two of the three modalities randomly and applied DCCA \cite{andrew2013deep} to obtain embeddings that are then concatenated and input to SVM.\newline
\textbf{4. DGCCA:}
We used the  publicly released DGCCA code \cite{andrew2013deep} to learn embeddings using a network architecture same as \textbf{dMCCA-1024-K} for different embedding dimensions, \textit{K}. The embeddings were then concatented and input to SVM.\newline
\textbf{5. MCCA:}
The projection vector space$\ \*V\in\mathbf{R}^{D\times K}$ was estimated on the training set as described in eqn. (4). The correlated components were estimated by:$\ \*Y^l=\*X^l\*V$ where the test set,$\ \*X^l\in\mathbf{R}^{T\times D}$ has \textit{T} samples and \textit{D}=784. The components are then concatenated and input to SVM.
\vspace{-0.1in}
\subsection{Performance Evaluation}
\begin{figure*}[t!]\label{fig:aff}
    \centering
    \includegraphics[width=0.7\textwidth]{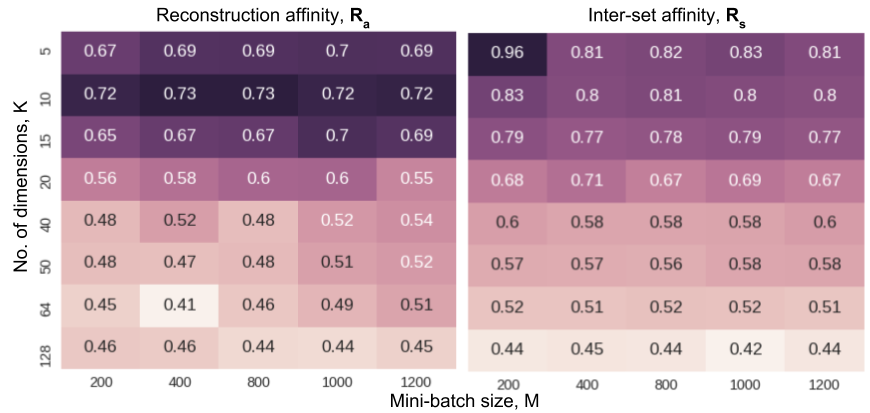}
    \vspace{-0.1in}
    \caption{{\small{\textbf{dMCCA-512-10} affinity measures for synthetic data}}}
\vspace{-0.2in}
\end{figure*}
The benefit of using synthetic data is that we can examine what the network learns when the generative process is known. 
Because components with equal ISC can be produced by arbitrary linear combination of the vectors in the corresponding subspace, we examined the \textit{normalized affinity measure} to compare dMCCA representation with the source signal. This was defined in \cite{soltanolkotabi2014robust} to estimate the angle between two subspaces. 
Let $\ \Hat{\*X}_s^l \in \mathbf{R}^{T\times K'}$ be the embedding for the source signal $\ \*X_s^l \in \mathbf{R}^{T\times K}$. The affinity between $\ \Hat{\*X}\text{ and }\*X$ can be estimated using the principal angles$\ \theta^{(\cdot)}$ as:
{\small
\begin{equation}
    \text{affinity}(\*X, \Hat{\*X}) \nonumber\\= \sqrt{\frac{\cos^2\theta^{(1)} + ... +\cos^2\theta^{\text{min}(K,K')} }{\text{min}(K,K')}}
\end{equation}}
where,$\ a \wedge b \equiv min(a,b) $ and the cosine of the principal angles $\ \theta$ are the singular values of the matrix $\ \*U^{\top}\*V$ where$\ \*U$ and $\ \*V$ are the orthonormal bases for $\ \*X \text{ and } \Hat{\*X}$ respectively. 

This measure of correlation between subspaces has been extensively used to compare distance between subspaces in the subspace clustering literature \cite{soltanolkotabi2014robust}. This measure is low when the principal angles are nearly orthogonal and has a maximum value equal to one when one of the subspaces is contained in the other. It can also compare subspaces of different dimensions.

We estimate two affinity measures: 1) \textit{reconstruction affinity}, $\ R_a$: average affinity between the reconstructed signal and the source signal across the$\ N$ modalities and 2) \textit{inter-set affinity}, $\ R_s$: average affinity between the different modalities of the reconstructed signal:
{\small
\begin{eqnarray}
    R_a = \frac{1}{N}\sum_{l=1}^{N}\text{aff}(\*X_s^l,\Hat{\*X}_s^l) \\
    R_s = \frac{2}{N(N-1)}\sum_{l=1}^{N}\sum_{{\substack{k=1 \\ l\neq k}}}^{N}\text{aff}(\Hat{\*X}_s^l,\Hat{\*X}_s^k) 
\end{eqnarray}}
\vspace{-0.1in}
\section{Results}\label{sec:results}
\textbf{Synthetic data results}:
We first analyzed the performance of dMCCA algorithm by varying embedding dimension and mini-batch size.
\textbf{Fig. 1} shows the reconstruction affinity measure ($\ R_a$ eqn. 11) and the inter-set affinity measure ($\ R_s$ eqn. 12).
Notice that the maximum$\ R_a$ is achieved for the embedding dimension of 10 (which is the number of correlated components used to generate the data) indicating that the dMCCA retains some notion of the ambient dimension for maximizing correlation between modalities.
The $\ R_s$ measure consistently decreased with increasing embedding dimension.
Because we estimate covariances in the loss function and use SGD with mini-batches for optimization, we also examined the performance with varying mini-batch size, \textit{M}. As shown in Fig. 1, $\ M>400$ gives consistent results. All our results were comparable for tanh activation.
\begin{figure}[]\label{fig:nmnist}
    \centering
    \includegraphics[width=0.45\textwidth]{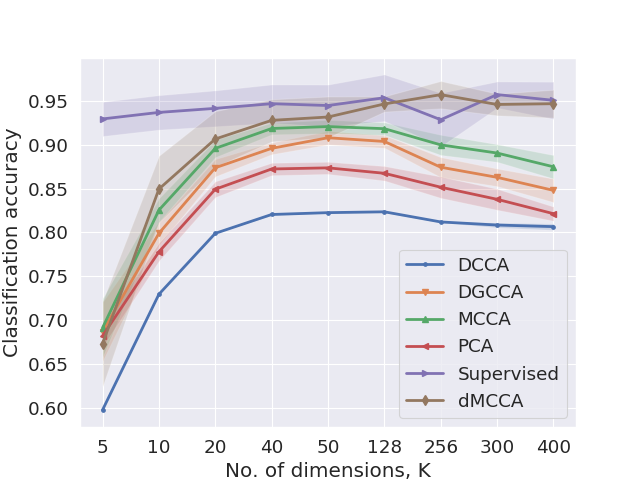}
    \vspace{-0.1in}
    \caption{Performance evaluation of dMCCA ($\ \bLozenge$)}
\end{figure}

Next, we compared the performance of our system with an empirical upper-bound obtained by training a DNN to reconstruct the ground-truth source signal. 
As shown in Table 1, and as expected -- the $\ R_a$ measure for our system (0.73) is lower than the supervised system (0.85). However, the affinity between the modalities for the embeddings from dMCCA (0.82) is higher than that of the supervised system (0.60). This is perhaps the benefit of modelling the between-set covariance over just minimizing the reconstruction error that is common to many deep representation learning methods, as well as DGCCA.
\begin{table}[t!]\label{tab:aff}
\centering
\caption{dMCCA vs. supervised method: Affinity measures}
\scalebox{0.8}{
\begin{tabular}{l|lll}
\toprule
Method                             & Activation & $\ R_a$                      & $\ R_s$                      \\
\midrule
\multirow{2}{*}{Supervised-512-10} & linear     & 0.85 $\ \pm$ 0.02 & 0.60 $\ \pm$ 0.03 \\
                                  & tanh       & 0.84 $\ \pm$ 0.01 & 0.45 $\ \pm$ 0.01 \\
\midrule                                   
\multirow{2}{*}{dMCCA-512-10}      & linear     & 0.73 $\ \pm$ 0.02 & 0.82 $\ \pm$ 0.01 \\
                                  & tanh       & 0.76 $\ \pm$ 0.01 & 0.78 $\ \pm$ 0.03 \\
\midrule                                   
Random baseline                    & --         & 0.05 $\ \pm$ 0.01 & 0.07 $\ \pm$ 0.03\\
\bottomrule
\end{tabular}}
\vspace{-0.2in}
\end{table}

\textbf{n-MNIST results: }We next evaluated our model to classify n-MNIST digits to assess dMCCA embeddings for classification task. Note that classifying noisy MNIST data is significantly difficult than the clean MNIST data \cite{basu2017learning}.
We compared the performance of all models by varying the embedding dimensions. 
10-fold CV accuracy averaged across mini-batch sizes and different partitions is shown in Fig. 2 (solid line). The standard deviation of accuracy is reflected by the error bars in Fig. 2.
While a supervised DNN trained end-to-end performed slightly better than our method, the performance was not statisticall different for $\ K>40$. 
PCA on the raw features yielded a best accuracy of about 87\% (red in Fig. 2) which was significantly greater than that of DCCA which only looked at two modalities.
CCA-based methods that used all three modalities outperformed DCCA method suggesting that multiple modalities can be leveraged more effectively to improve the performance. 

Finally, our model outperformed the linear features from MCCA and other CCA methods suggesting the benefit of using deep learning to model the non-linear relationships.
A key factor here is the embedding dimension$\ K$. In real-world data, this number is not known. But, in practice, it can be tuned on a validation set. 
Increasing the dimension size further did not improve classification performance. 
We also performed clustering on the dMCCA embeddings to assess linear separability (NMI=0.72, completeness=0.67). This suggests that dMCCA method can also be used for unsupervised learning.
\vspace{-0.1in}
\section{Discussion and Conclusions}\label{sec:conc}

This ratio of variances used in our loss is also referred to as intraclass correlation coefficient \cite{bartko1966intraclass} and is a widely-used measure of reliability in test-retest literature.
Hence, multiple modalities can be viewed as analogous to repeated, albeit noisy measurements of the underlying signal.
For classifying noisy MNIST data, we show that our model not only outperforms other CCA-based approaches but competes well with supervised learning models.
For future work, we are working to include dMCCA within an autoencoder framework, and explore interpretability of features using MCCA. We have released the code and results on more real-world examples at https://github.com/usc-sail/mica-deep-mcca

\bibliographystyle{unsrt}
\bibliography{references}

\end{document}